\documentclass[runningheads]{llncs}
\usepackage{graphicx}
\usepackage{float}

\begin{document}
\title{Law to Binary Tree - An Formal Interpretation of Legal Natural Language}
%
%
\author{Ha-Thanh Nguyen\textsuperscript{1,}\thanks{Corresponding: nguyenhathanh@nii.ac.jp.} \and 
Vu Tran\textsuperscript{2} \and
Ngoc-Cam Le\textsuperscript{3} \and 
Thi-Thuy Le\textsuperscript{4} \and 
\\Quang-Huy Nguyen\textsuperscript{5} \and 
Le-Minh Nguyen\textsuperscript{6} 
\and Ken Satoh\textsuperscript{1}
}
%
\pagestyle{empty}
%
\institute{National Institute of Informatics, Tokyo, Japan 
\and The Institute of Statistical Mathematics (ISM), Tokyo, Japan 
\and Vietnam Judicial Academy, Hanoi, Vietnam 
\and Hanoi Law University, Hanoi, Vietnam 
\and VINASECO JSC, Hanoi, Vietnam 
\and Japan Advanced Institute of Science and Technology, Ishikawa, Japan}
\maketitle              
\begin{abstract}
Knowledge representation and reasoning in law are essential to facilitate the automation of legal analysis and decision-making tasks. 
In this paper, we propose a new approach based on legal science, specifically legal taxonomy, for representing and reasoning with legal documents. 
Our approach interprets the regulations in legal documents as binary trees, 
which facilitates legal reasoning systems to make decisions and resolve logical contradictions. 
The advantages of this approach are twofold. 
First, legal reasoning can be performed on the basis of the binary tree representation of the regulations. 
Second, the binary tree representation of the regulations is more understandable than the existing sentence-based representations.
We provide an example of how our approach can be used to interpret the regulations in a legal document.
\keywords{legal science \and legal representation \and  binary tree \and interpretation}
\end{abstract}
\section{Introduction}

In recent years, artificial intelligence technology has been widely used in the legal field to facilitate legal analysis and decision-making tasks~\cite{araszkiewicz2021identification,xu2021accounting,nguyen2021few}. 
Although legal documents usually describe the regulations in natural language, there is an ordered and logical system based on legal doctrines and theoretical legal issues behind the words. If it is only based on natural language processing without focusing on the systematic structure of legal documents, an artificial intelligence system is likely to give inaccurate or meaningless outcomes for the tasks which require logical analysis, or systematic legal knowledge. 
Therefore, using a formal logical representation to represent the regulations in legal documents is a promising approach. 
In more detail, jurists can express systematic legal knowledge through formal logical representation, while computers can easily process formal logical representations and ``understand'' the legal logic made by jurists. Thereby, artificial intelligence can simulate and learn how lawyers analyze certain legal documents. In other words, through formal logical representation, artificial intelligence and jurists can communicate logical problems with each other.

Many approaches have been proposed to represent the regulations in legal documents as logic formulas. 
The legal rules can be represented in the form of a structured representation~\cite{nguyen2017knowledge}, a rule-based representation~\cite{satoh2010proleg} or a graph-based representation~\cite{sovrano2020legal}. 
These approaches share the same general idea of representing the regulations as formal rules. 
We find that the main challenges for bringing the work to practical applications are not only in the technology of representing legal norms but also in the method of developing the technology with the involvement of legal experts.
It costs a huge amount of time for lawyers to understand the logical expression and then to produce the logical formulas from the legal documents.

In this paper, we propose a new approach for representing and reasoning with legal documents. 
Our approach interprets the regulations in legal documents as binary trees and employs a legal reasoning system to resolve logical contradictions. 
We prove that our approach can be used to perform legal reasoning on the basis of the binary tree representation of the regulations. 
The biggest advantage of our approach is that the binary tree representation is easy to understand and to produce by jurists. 
As a result, our approach can be used to develop an automatic legal reasoning system that is understandable by jurists. 

\section{Preliminaries}
The legal system of a country is composed of vast legal norms. Taxonomy plays an essential role in the arrangement of these legal regulations. Thanks to taxonomy, jurists think logically about legal problems by sorting legal rules, deciding them into categories, and generalizing them into fields of law \cite{Sherwin2009}. Therefore, although the legal system consists of massive legal norms, it is still orderly, logical, and coherent. That is why jurists often use a logic diagram that we call a latent tree structure to analyze legal regulations and deal with legal documents. In this way, it is easier for them to see the relationship and hierarchy between the elements of the legal system. The word ``latent" means that there is a hidden tree structure behind the legal system, and the tree structure is generated by the relationship between the legal rules and the fields of law. 

A latent tree structure often consists of two parts: 
(i) \textbf{The root} presents legal taxonomies which are widely analyzed and acknowledged in legal science. This part may have similarities between the laws of countries in the same legal traditions, i.e., civil law, common law, religious law, etc.; and
(ii) \textbf{The rest} shows the contents of current legal norms stuck with each legal taxonomy mentioned above. Depending on the policies of each state or nation, the part may be differences between the laws of each country, even though they share the same legal traditions.



\section{Proposed Approach}
As mentioned in the previous section, the latent tree structure is an effective tool for jurists to deal with legal documents. The latent tree can be drafted manually on paper or in jurists' minds without any standard. If there is a method to build a latent tree structure from a legal document automatically, it will make jurists' work more efficient and accurate. In addition, we can build an AI system that could automatically interpret legal rules in a manner similar to how lawyers and judges would.

Humans can use their common sense and legal experience to prioritize the rules. However, this is difficult for the machine or even lay people.
For example, when checking a contract, if the agreement contains both particular rules and  general rules, the particular rules take priority over the general rules. 
Another example is that before checking the contract, we need to check the parties to determine if they are qualified. Are they eighteens or older? Do they have the mental capacity to understand the contract? Which qualiﬁcations should take priority when there is a conflict between them? When the contract contains many qualifications, it increases the complexity of priority checking.

We propose a new way to resolve the problem of priority relationships between rules in latent trees. The idea is to construct a binary tree structure, in other words, convert an ordinary latent tree into a binary tree. In a binary tree, there are only two child nodes for each parent node. The position of nodes can tell which rule needs to be considered first. For example, the root is always the first node, and based on the result of the root we can go to the left or right node. When we get to the leaf, we can get the final result. This representation can be used by the machine to automatically resolve the conflict between the rules and determine the priority.

\section{Case Study}

In this section, we use a part of legal norms on inheritance under wills with a focus on the regulations related to testators to demonstrate how to convert the latent tree into a binary tree. 

\noindent{\textbf{Step 1: Drawing the latent tree}}

\noindent Please note that, in practice, jurists will draw this tree on an AI system based on (i) technically fixed standards; and (ii) their systematic legal knowledge of relevant legal norms.

First of all, the root of the latent tree structure can be drawn based on legal taxonomies of laws on inheritance under wills. 

From a legal point of view, the legal norm is created to govern the respective legal relations. Jurists can classify a legal relation into three elements: \textbf{a subject}, \textbf{an object}, and \textbf{contents} of legal relation\cite{araszkiewicz2021identification,Khoshimov2020}. Besides, when analyzing and assessing a legal relation, it is essential to consider its creation, change, and termination.  
Thus, first of all, we draw the root of the latent tree structure includes four nodes: \textbf{subjects} in the inheritance relation under wills, the \textbf{object} of the inheritance relation under wills, \textbf{contents} of the inheritance relation under wills, and \textbf{the creation, change, and termination} of the inheritance relation under wills. Then, we break down each element separately in detail as follows (Figure 1):

(i) \textbf{Subjects}: In legal science, a subject (a legal person) includes two kinds: \textbf{a natural person} (human person, sometimes also a physical person), and \textbf{a legal entity} (non-human person) which is a body that can function legally, sue or be sued, and make decisions through agents such as association, corporation, partnership \cite{garner2014black,web:LIIlegalperson}. Thus, main subjects in the inheritance under wills may include the testator, heirs, administrators of estates, and witnesses. 

(ii) \textbf{The object}: In legal science, object is what the subjects want to achieve when entering into a legal relation, or the aim or purpose of legal norms which govern the respective legal relation \cite{garner2014black,GTUc2017}. So, estates are the object of the inheritance under wills relation.

(iii) \textbf{Contents}:  In legal science, contents of a legal relation are legal rights, and obligations of the relevant subject(s)\cite{araszkiewicz2021identification,Khoshimov2020,GTUc2017}. Hence, the contents of the inheritance relation under wills are rights as well as obligations of each relevant subject, i.e., the testator, heirs, administrators of estates, and witnesses.

(iv) \textbf{the creation, change and termination} of the inheritance relation under wills.

\begin{figure}[H]
    \centering
    \includegraphics[width=\textwidth]{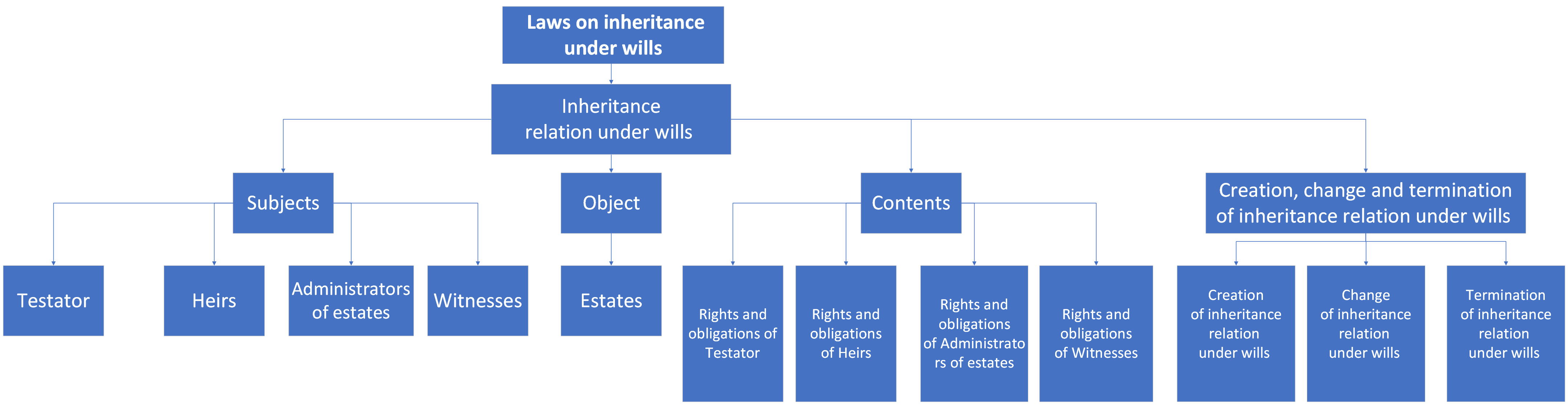}
    \caption{The root of the latent tree structure based on legal taxonomies of laws on inheritance under wills}
    \label{latent_tree_1}
\end{figure}

The laws of inheritance under the wills of many countries such as Vietnam\cite{Vietnamesecivilcode2015}, China\cite{chinacivilcode2020}, Germany\cite{Germancivilcode2002}, France\cite{Francecivilcode2016} etc. should be classified the same taxonomy mentioned above. 
Figure \ref{latent_tree_1} shows the root of the latent tree structure based on legal taxonomies of laws on inheritance under wills.

\textbf{Next}, we draw the rest of the latent tree structure which shows the contents of current legal norms on the inheritance relation under wills stuck with each legal taxonomy mentioned above. As mentioned above, this part of the tree structure is different when representing the legal norms of different countries due to differences in the practical laws of each country. After the task, we have a complete latent tree structure on the inheritance relation under wills.
For illustration purposes, we only demonstrate legal norms on testators in two countries: Vietnam, and China.

\begin{figure}
    \centering
    \includegraphics[width=\textwidth]{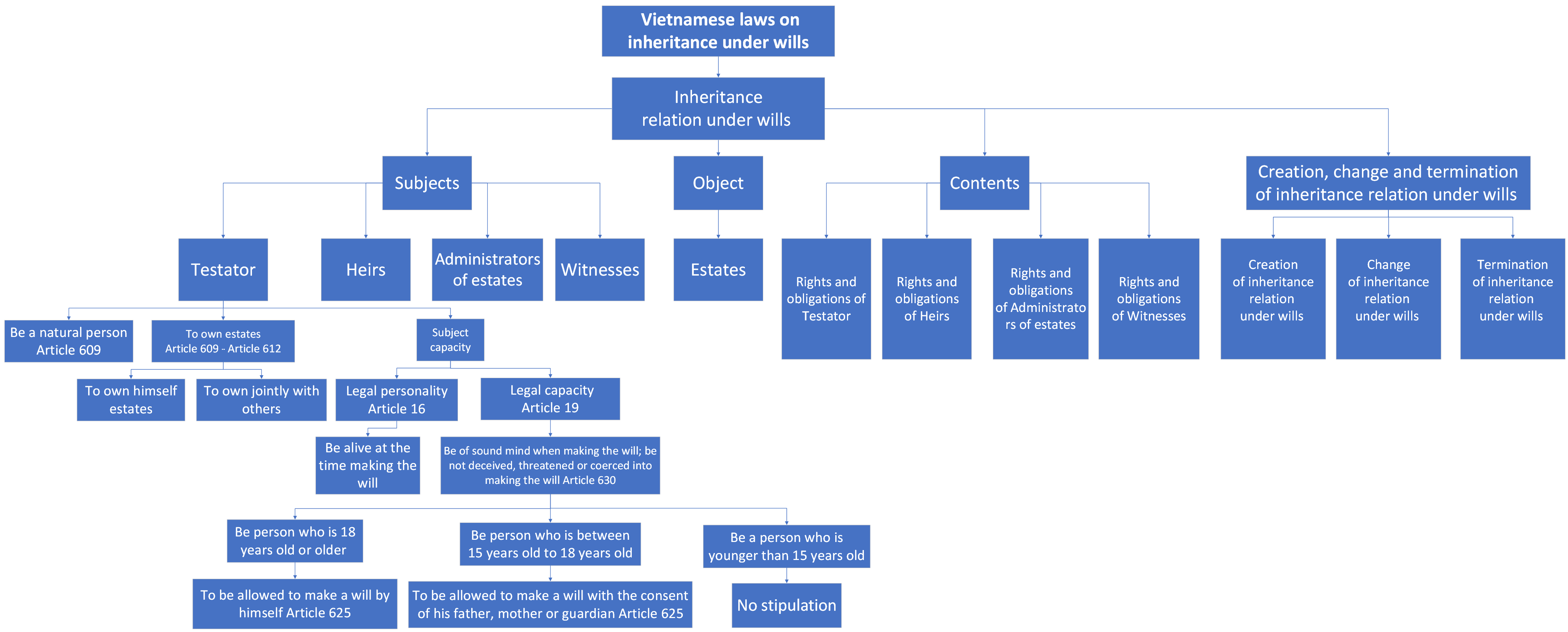}
    \caption{The latent tree structure based on Vietnamese laws on inheritance under wills}
    \label{Vietnamese_latent_tree}
\end{figure}

\begin{figure}
    \centering
    \includegraphics[width=\textwidth]{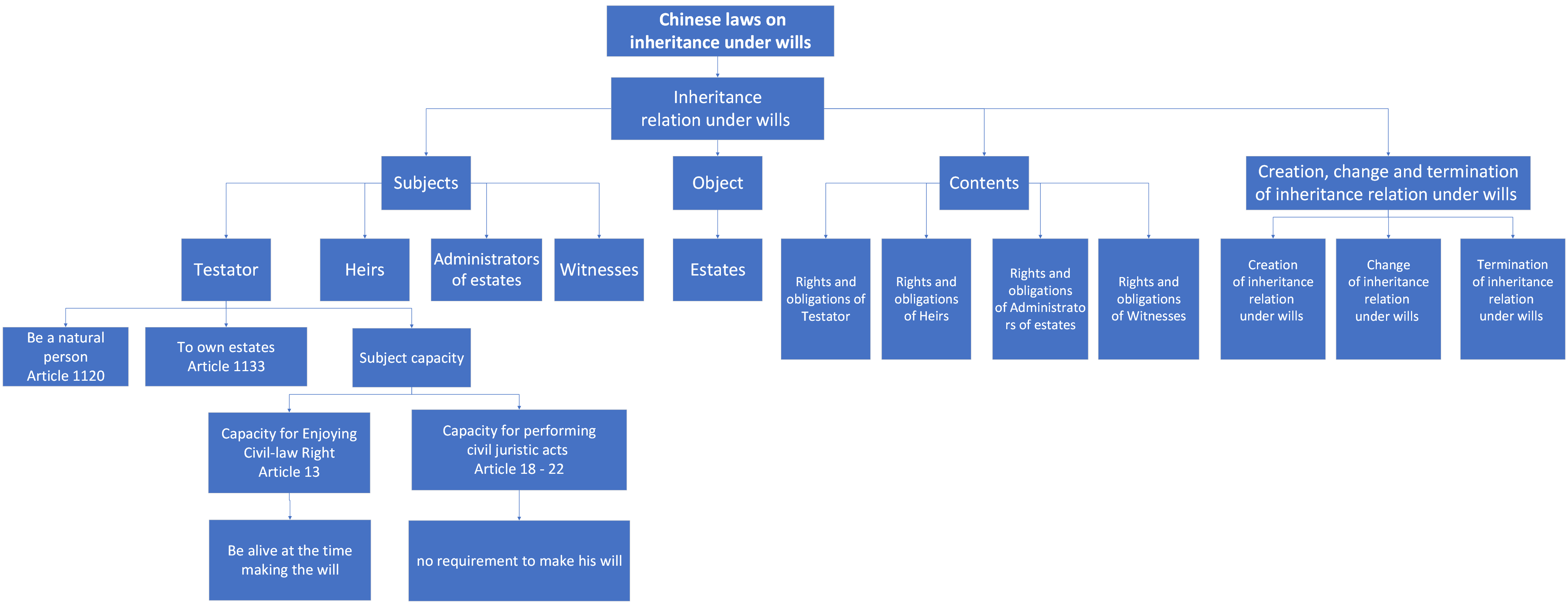}
    \caption{The latent tree structure based on Chinese laws on inheritance under wills}
    \label{Chinese_latent_tree}
\end{figure}

Figure \ref{Vietnamese_latent_tree} and \ref{Chinese_latent_tree} show latent trees of Vietnamese and Chinese legal norms. We can see that both trees have a similar structure. In general, legal norms for this case are categorized into four groups: subjects,  objects, contents, and the problem of creation, change, and termination of inheritance relation under wills. The trees are slightly different in terms of gestation and subject capacity. These differences may lead to big differences in legal consequences. For example, in Vietnam law, the legal capacity of the testator is divided into three cases: under 15 years of age, from 15 to 18 years of age, and over 18 years of age. The age of the testator is one of the important elements that determine the status of inheritance relation under wills.

\noindent\textbf{Step 2: Converting the latent tree into the binary tree}

\begin{figure}
    \centering
    \includegraphics[width=\textwidth]{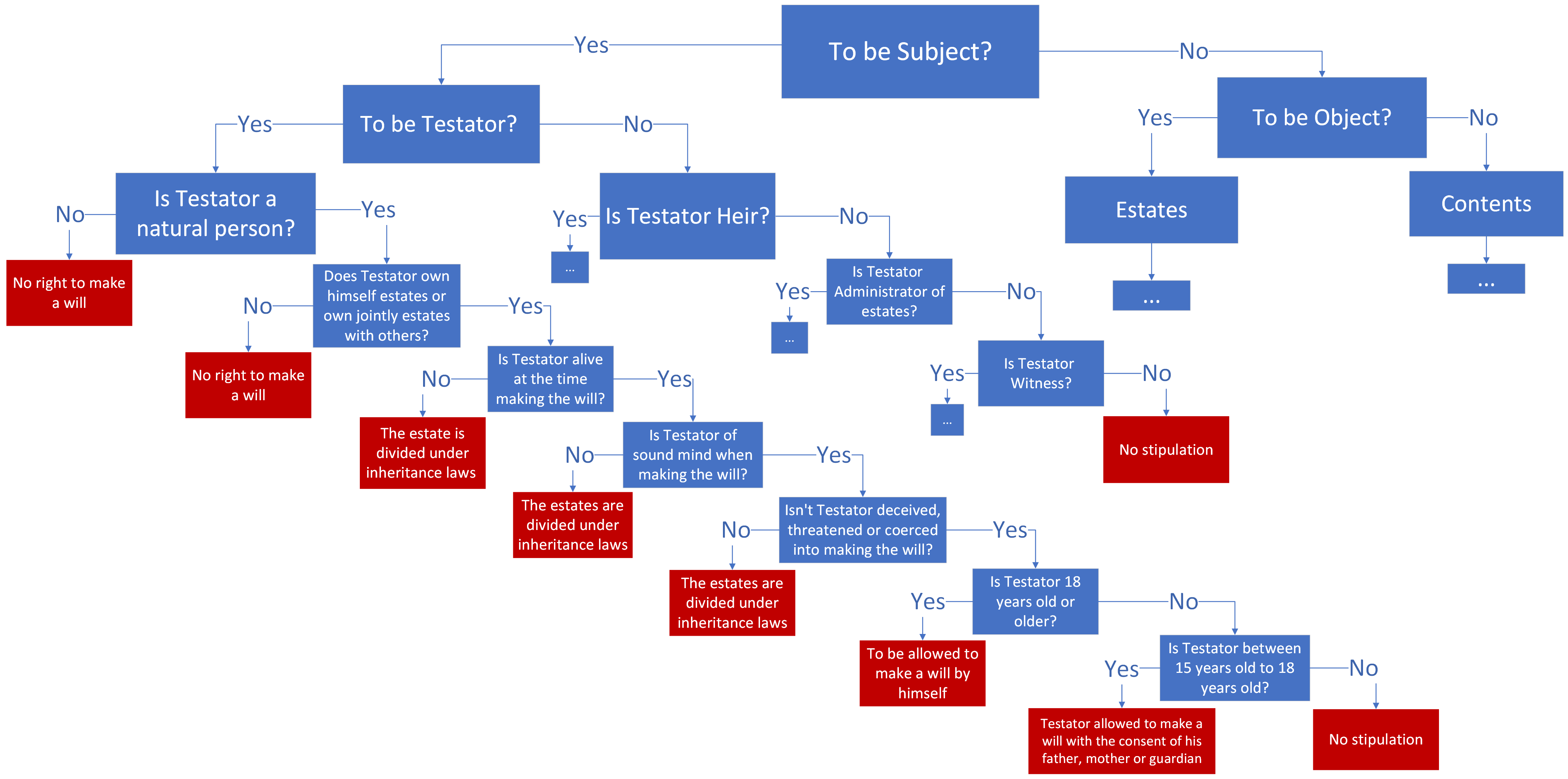}
    \caption{The binary tree structure based on Vietnamese laws on inheritance under wills}
    \label{Vietnamese_binary_tree}
\end{figure}

\noindent An example of the Vietnamese binary tree is shown in Figure \ref{Vietnamese_binary_tree}. 
This tree has a structure similar to a decision tree, with two branching arrows standing for Yes and No. 
The leaf nodes are the legal consequences of the interpretation. 
For example, in the case of Vietnamese law, if the testator is not a natural person, 
then there is no right to make a will.
This approach allows us to reason about legal conclusions automatically. 
Besides, we can also utilize the game theory or tree-based algorithm to make efficient and effective legal decision-making applications with this structure.

With this binary tree structure, we create a deterministic and complete reasoning system,
in which the legal consequences are determined by the input features. 
This reasoning system is not only suitable for legal decision-making applications but also an effective tool for analyzing legal regulations.
For example, if there is no logical way to arrange the nodes, 
this can be considered a contradiction in the legal regulation.
In addition, this structure can be used to frame the reasoning process of black-box models when learning from data.

\section{Discussion and Conclusion}
The proposed method for building the binary tree from the legal document is a way to understand the structure of legal concepts and regulations and resolve conflicts between rules automatically. 
This is important because it can make the machine automatically interpret legal rules in a manner that is similar to how lawyers and judges would. 
It will also make lawyers' work more efficient and accurate.
In this paper, we also introduce a case study in which we can see the application of the proposed method to inheritance analysis to demonstrate our idea better.
Note that the example tree presented in this paper is only one possible way to construct the binary tree. 
The tree can be customized to fit the different needs of the particular legal problem. 
This representation can be used to validate the outcome and reasoning ability of black-box machine learning models.
In future works, we will involve this proposal in our systems, research activities, legal activities, and competitions.

\bibliographystyle{splncs04}
\bibliography{ref.bib}
\end{document}